\documentclass[conference]{IEEEtran}
\IEEEoverridecommandlockouts
\usepackage{cite}
\usepackage{amsmath,amssymb,amsfonts}
\usepackage{algorithmic}
\usepackage{graphicx}
\usepackage{textcomp}
\usepackage{changepage}
\usepackage{multirow}
\usepackage{booktabs}
\usepackage{xcolor}
\def\BibTeX{{\rm B\kern-.05em{\sc i\kern-.025em b}\kern-.08em
    T\kern-.1667em\lower.7ex\hbox{E}\kern-.125emX}}
\begin{document}

\title{SigScatNet: A Siamese + Scattering based Deep Learning Approach for Signature Forgery Detection and Similarity Assessment}

\author{\IEEEauthorblockN{Anmol Chokshi*, Vansh Jain*, Rajas Bhope*, Sudhir Dhage*}

\IEEEauthorblockA{Sardar Patel Institute of Technology, All authors contributed equally*}}

\maketitle

\begin{abstract}
The surge in counterfeit signatures has inflicted widespread inconveniences and formidable challenges for both individuals and organizations. This groundbreaking research paper introduces SigScatNet, an innovative solution to combat this issue by harnessing the potential of a Siamese deep learning network, bolstered by Scattering wavelets, to detect signature forgery and assess signature similarity. The Siamese Network empowers us to ascertain the authenticity of signatures through a comprehensive similarity index, enabling precise validation and comparison. Remarkably, the integration of Scattering wavelets endows our model with exceptional efficiency, rendering it light enough to operate seamlessly on cost-effective hardware systems. To validate the efficacy of our approach, extensive experimentation was conducted on two open-sourced datasets: the ICDAR SigComp Dutch dataset and the CEDAR dataset. The experimental results demonstrate the practicality and resounding success of our proposed SigScatNet, yielding an unparalleled Equal Error Rate of 3.689\% with the ICDAR SigComp Dutch dataset and an astonishing 0.0578\% with the CEDAR dataset. Through the implementation of SigScatNet, our research spearheads a new state-of-the-art in signature analysis in terms of EER scores and computational efficiency, offering an advanced and accessible solution for detecting forgery and quantifying signature similarities. By employing cutting-edge Siamese deep learning and Scattering wavelets, we provide a robust framework that paves the way for secure and efficient signature verification systems.
\end{abstract}


\section{Introduction}

With the rising digitization of papers and transactions, signature forgery has become a major concern. Traditional manual signature verification procedures are frequently subjective, time-consuming, and error-prone. As a result, there is a growing demand for automated systems and technologies that successfully identify and prevent signature fraud using computer vision, machine learning, and biometric identification approaches. A number of papers from the research community have worked on using machine learning for detecting forgery in signatures. While their work has provided useful insights into their methodology, there are a few areas where improvements can be made in order to make this system more robust.

Most papers from the research community use a basic CNN network to identify if the signature is fraudulent or not. For employing this, they use a dataset containing signature images and then apply pre-processing techniques and feed this into the network. Now for testing, they provide an image that is then compared with the images from the dataset. The output of this testing comes out as a prediction which is then compared with a threshold to verify if the image is a forgery or not. While this system works, it has its drawbacks. Firstly, signatures can vary naturally due to factors like the signer's attitude, exhaustion, or writing pen type, making it difficult to set a general threshold that successfully distinguishes authentic from faked signatures. As a result, this strategy may produce more false negatives or false positives, compromising the system's reliability. Furthermore, the success of CNN-based signature forgery detection is largely dependent on the quality and variety of the training dataset. Building a comprehensive and representative dataset is a difficult endeavor that necessitates great effort in collecting a diverse range of authentic and counterfeit signatures. Additionally, CNN-based signature forgery detection systems often struggle to account for sophisticated forgery techniques, such as skilled imitations or forged signatures created using advanced tools. These techniques can potentially deceive the CNN network, leading to incorrect predictions and reduced accuracy. Furthermore, the reliance on pre-processing techniques introduces additional complexity and computational overhead, impacting the overall efficiency of the system.

The SigScatNet employs a more complex form of CNN. The Siamese model with Scattering wavelets counteracts the limitations discussed by leveraging its unique architecture and comparison-based approach. The Siamese model, as opposed to a simple CNN’s basic threshold-based decision-making, learns a similarity measure that allows for a more flexible and comprehensive evaluation of authentic or forged signatures. The Siamese model can reflect the intrinsic complexity and diversity in signatures by examining pairwise interactions between signatures, minimizing false negatives and false positives. Furthermore, the model achieves lightweight efficiency while preserving critical spatial and temporal information by using Scattering wavelets. This improves performance in capturing dynamic elements of signatures and adjusting to evolving forgery tactics, overcoming the limits of standard CNN-based methods.

\section{Related Work}
The identification of signature forgery using machine learning techniques has seen significant advancements, as evidenced by several research papers. Convolutional neural networks (CNNs) combined with image processing techniques have been commonly employed in many papers.~\cite{1} developed a deep learning-based off-line signature verification system using a Convolutional Neural Network (CNN) and a novel method for extracting local features to distinguish between genuine and forged signatures, with potential applications in organizations with a limited number of individuals for training and future forgery detection.~\cite{2} proposes a system for signature verification and forgery detection wherein this paper uses CNN and deep learning to extract unique features from pre-processed signatures, which are then compared to the signatures stored in the system to determine whether the signature is real or fake. The system was tested on a dataset containing 750 signatures from 150 individuals, and the results showcased how the proposed system works and then stated that their method is more accurate and faster in the detection of forged signatures than a trained professional.~\cite{3} developed a solution using Convolutional Neural Network (CNN) to detect forgery in handwritten signatures by training the model on a dataset of signatures and predicting the authenticity of provided signatures.~\cite{4} developed an interactive software implementation of signature verification that includes both general learning from a population of signatures and special learning from multiple samples of an individual's signature, resulting in improved accuracy.~\cite{5} proposes a signature verification system that uses pre-processing techniques such as RGB to Gray Scale Conversion, image binarization, and Gaussian blurring. The system uses contour and SIFT features for feature extraction and SVM and K-Means algorithms for model training.

Siamese networks have gained attention in recent research. In~\cite{6}, the methodology used involves the CEDAR dataset. The signatures are processed using various techniques such as median filter, Otsu method, and morphological operations. The Siamese neural network is then trained and tested on this dataset, resulting in high accuracy rates.~\cite{7} conducted offline signature verification using a Siamese neural network based on one-shot learning, which achieved successful classification with fewer signature images. By utilizing the one-shot learning method, the model distinguished between genuine and forged signatures without the need for a large amount of labeled data. Experimental results on signature datasets demonstrated across 4NSigComp2012, SigComp2011, 4NSigComp2010, and BHsig260 datasets, showcasing the effectiveness of their proposed approach.~\cite{8} uses a Siamese Neural Network (SNN) with a contrastive loss function and robust embedding to verify genuine and forged offline signatures. The experiments were conducted on challenging datasets like MCYT-75, GPDS, and CEDAR, and the proposed method outperformed other state-of-the-art methods in terms of accuracy. Their results showcased that the use of inter-quartile range and median absolute deviation measures in the embedding vector increases the accuracy of the algorithm for GPDS and MCYT-75 databases relating to the signature forgery domain. However, existing approaches often face limitations in handling natural variations that contribute to increased false positives and false negatives, or they require computationally expensive networks with high hardware requirements.

This research paper aims to address these limitations by striking a balance between the robustness of the system and the canonical form in signature forgery detection. The objective is to develop an approach that effectively captures natural factors influencing signatures while maintaining computational efficiency. By employing a Siamese network and incorporating novel techniques such as Scattering wavelets, the SigScatNet aims to achieve accurate and lightweight signature forgery detection. This would contribute to enhancing the reliability and effectiveness of signature verification systems, improving overall security and trust in various domains where signatures play a crucial role.

\section{Proposed Methodology}

\subsection{SigScatNet Architecture}

\begin{figure*}[ht]
  \centering
  \includegraphics[width=1\linewidth]{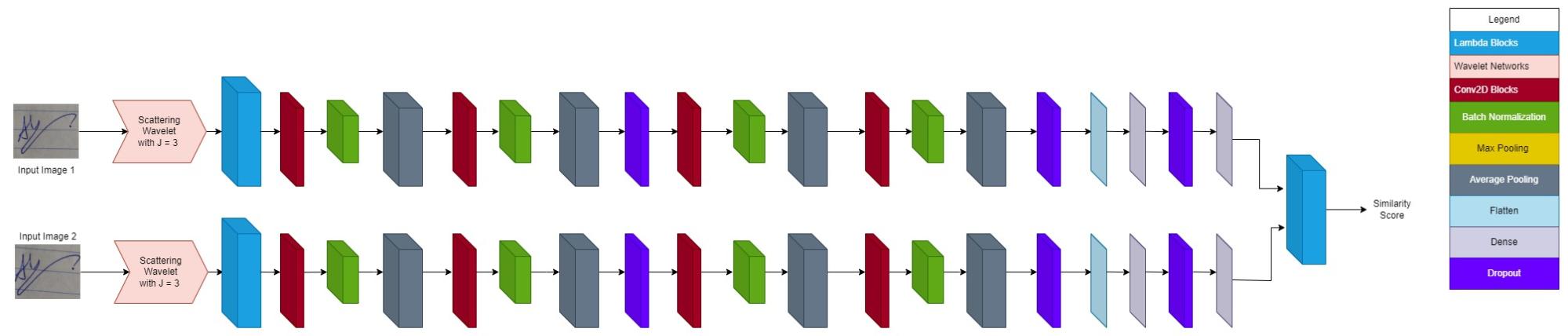} 
  \caption{SigScatNet Architecture}
  \label{fig:architecture}
\end{figure*}

Fig. 1 illustrates the SigScatNet architecture, showcasing the integration of Scattering wavelets and the Siamese network. This architecture forms the core of SigScatNet, enabling the robust detection of signature forgery and precise assessment of signature similarity.

The SigScatNet architecture can be divided into two essential parts: the Scattering wavelet component and the Siamese network component.

\subsubsection{Scattering Wavelets}
Scattering wavelets are an effective technique for extracting useful characteristics from signals or images in signal processing and image analysis. Both local and global information is captured in a multiscale, translation-invariant representation provided by this method. The wavelet transform, which breaks down a signal into several frequency components at various scales, serves as the foundation for the scattering transform. The scattering transform is better suited for tasks like pattern identification and signal classification since it keeps the information about phase and magnitude in contrast to conventional wavelet transforms. The scattering transform works via cascading wavelet transformations and modulus non-linearities. A series of wavelet filters are convolved with the input signal or image at various scales. Insight into local frequency is captured by the resulting wavelet coefficients. The magnitude data is then captured by applying the modulus non-linearity to the wavelet coefficients. A depiction of multiscale scattering is created by repeating this procedure at several scales.

The scattering transform is incredibly helpful when it comes to signature analysis. The level of depth in a signature can range from broad strokes to finer details. We can record these various informational scales by applying the scattering transform on signature images. The signatures' global and low-frequency components, as represented by the coefficients determined from lower scales, are what give the signatures their overall shape and structure. On the other hand, finer details and high-frequency components, such as minor texture patterns and variation in stroke, are captured by the coefficients from higher scales. We consider a handwritten signature as an illustration in Fig. 2. The scattering transform produces a scattering representation with several scales of coefficients when applied to the signature image. Lower scale coefficients encode the signature's overall shape and structure, encapsulating its general properties. The factors pick up finer features at larger scales, including differences in the thickness of the signature's strokes, their curvature, and their texture patterns. In this process, the selection of scales and scattering factors is critical. We may regulate the level of detail recorded in the scattering representation by choosing the right scales. Finer features can be captured at higher scales, however doing so may also result in an increase in noise or artifacts. It's crucial to find a balance and pick scales that accurately depict the pertinent distinctive traits without obscuring the representation. The scattering transform is further influenced by the choice of wavelet filters, their number per scale, and their sizes. These variables affect how the transform behaves and how many features are extracted. These parameters need to be carefully thought out in order to guarantee ideal feature extraction and discriminative power for signature analysis jobs.

\begin{figure*}[ht]
  \centering
  \includegraphics[width=0.8\linewidth]{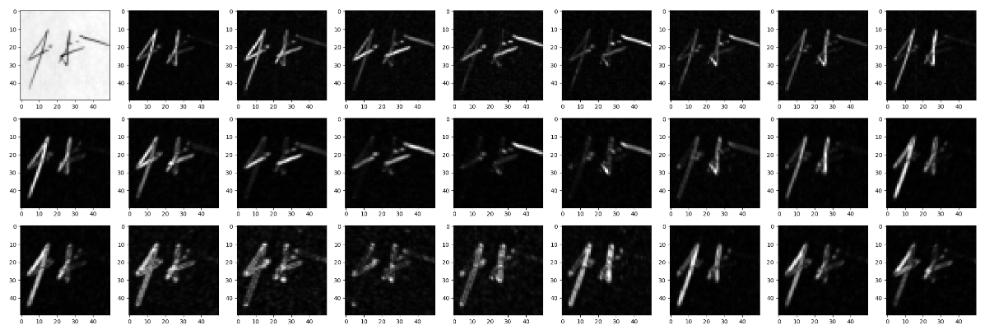}
  \includegraphics[width=0.8\linewidth]{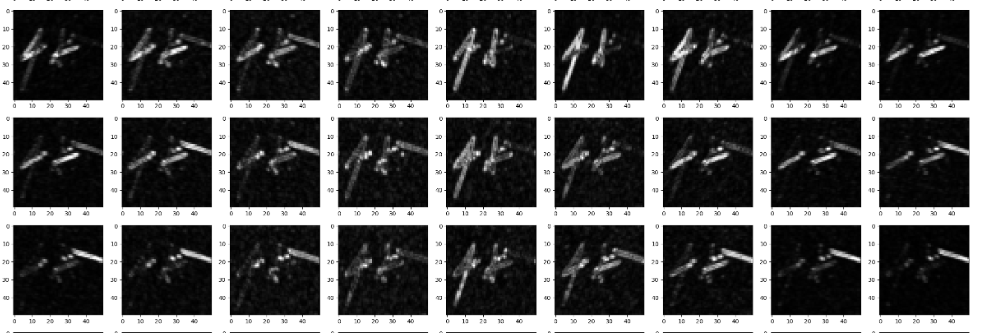}
  \includegraphics[width=0.8\linewidth]{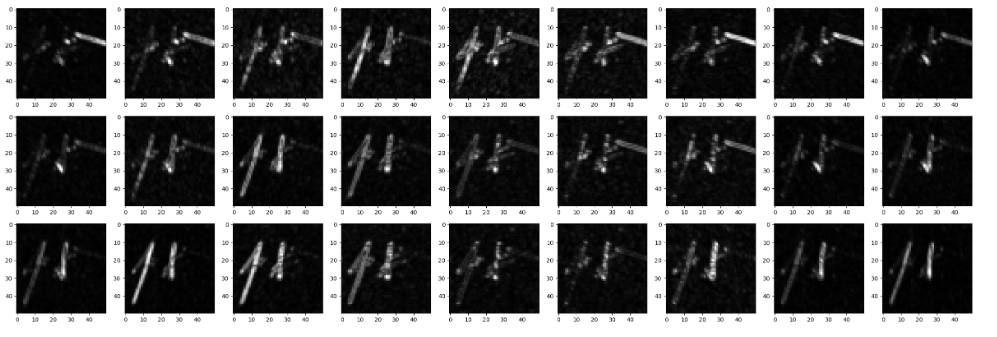}
  \caption{Signatures at J=2 on performing Scattering Wavelet Operation.}
  \label{fig:scattering}
\end{figure*}

\subsubsection{Siamese Model}
The Siamese model is a potent method for comparing two signatures to assess their similarity. In our scenario, the Siamese model is used to compare a signature under scrutiny with a base or reference signature, allowing us to determine whether the signature is genuine or not. The Siamese model comprises of two identical neural networks with the same architecture and weights. Each network receives an input signature, processes it through a series of convolutional and fully connected layers to extract relevant features. The Conv2D layers in the model use filters with the following configurations: 16 filters in the first two layers and 32 filters in the last two layers. The kernel size for all Conv2D blocks is set to \(3\times3\), and the activation function used is ReLU. The features from the two networks are concatenated and passed into a similarity function.

The model takes an input image of size \(180 \times 300 \times 3\). Before the scattering network's operation, the image is converted to grayscale. It then goes through the convolutional blocks, resulting in an output array of size 128, which represents the signature's features. A similarity score is generated by comparing the feature embeddings of the two signatures using these 128-sized arrays. One commonly used similarity metric is the Euclidean distance, also known as the L2 distance.

The Euclidean distance \(D\) between the feature embeddings of input image 1 and input image 2 is computed as seen in equation (5), where \(n\) denotes the number of features (128 in our case):

\[
D = \sqrt{\sum_{i=1}^{n} (x_{i} - x_{i})^2}                     \hspace{30pt}\text{(5)}
\]

This Euclidean distance is normalized between 0 and 1, where a lower \(D\) denotes a higher degree of similarity between the signatures.

The Siamese model is trained using a triplet loss function during the training phase~\cite{9}. The triplet loss function is a distance-based loss function designed to learn embeddings by minimizing the distance between the anchor and positive samples while maximizing the distance between the anchor and negative samples. The triplet loss function plays a crucial role in learning discriminative embeddings by encouraging the model to bring genuine signatures (positive samples) closer to the anchor signature and push forged signatures (negative samples) farther away from the anchor. This enables effective differentiation between genuine and forged signatures based on their feature representations.

We utilize the trained Siamese model to determine a signature's similarity score to the reference signature for validation. If the similarity score exceeds a predefined threshold, the questioned signature is considered authentic; otherwise, it is labeled as a forgery.

The formula for the Triplet loss is shown in equation (6):
    
\[
L = \max(d(a,p) - d(a,n) + \text{$margin$},0)
\hspace{30pt} \text{(6)}
\]
Where $L$ represents the Triplet loss, $d(a,p)$ represents the distance between the anchor sample $a$ and the positive sample $p$, $d(a,n)$ represents the distance between the anchor sample $a$ and the negative sample $n$. The $margin$ is a hyperparameter that specifies a margin or threshold value, $max(x,0)$ ensures that the loss is non-negative, as the loss is only calculated when the positive distance plus the margin is greater than the negative distance. The model was trained with a learning rate of 0.0005, a batch size of 32, and it underwent 100 epochs of training.

\subsection{Dataset}
In our study, we addressed the task of signature verification and forgery detection using two different datasets: the ICDAR 2011 SigComp Dutch Signature Dataset and the CEDAR Dataset.

There are two folders in the ICDAR 2011 SigComp Dutch Signature Dataset: "train" and "test." The "train" folder contains samples from 64 people, each of whom has at least 24 genuine and 8 forged signatures. This dataset serves as the primary training set for our model. The "test" folder, on the other hand, contains samples from 42 different individuals, each with a minimum of 12 authentic and 12 forged signatures. We use this dataset to evaluate the effectiveness and generalizability of our trained model.

The CEDAR dataset comprises 1320 forgeries, with 24 forgeries for each of the 55 writers. Additionally, it contains 1320 genuine signatures, with 24 original signatures from the same 55 writers. Therefore, for each person, we have access to 24 original signatures and 24 forgeries. Each image is resized to 300x180 pixels before being fed into the model.

Out of the 55 individuals, 40 are utilized to train the model, while the remaining 15 people are reserved for testing. This approach ensures that the trained model is tested on entirely new signatures that it has not encountered during the training phase.

\section{Experiments and Results}
The evaluation of our SigScatNet model on the ICDAR SigComp Dutch dataset and the CEDAR dataset resulted in impressive performance metrics. The Receiver Operating Characteristic (ROC) curve, which plots True Positive Rate (TPR) against False Positive Rate (FPR), exhibited an exceptional Area Under the Curve (AUC) of 99.6\% on the ICDAR SigComp Dutch dataset and 99.9\% on the CEDAR dataset. This indicates the model's ability to achieve high true positive rates while maintaining low false positive rates, a crucial aspect for accurate signature forgery detection. Additionally, the Precision-Recall (PR) curve demonstrated an outstanding Average Precision (AUPR) of 99.4\% on the ICDAR SigComp Dutch dataset and 99.9\% on the CEDAR dataset, further reinforcing the model's effectiveness in correctly identifying genuine and forged signature pairs. These results validate the discriminative power of our SigScatNet model and its robustness in capturing signature characteristics. Fig. 3 and Fig. 4  showcases the ROC and PR curve on the two datasets.

\begin{figure}[ht]
  \centering
  \includegraphics[width=1\linewidth]{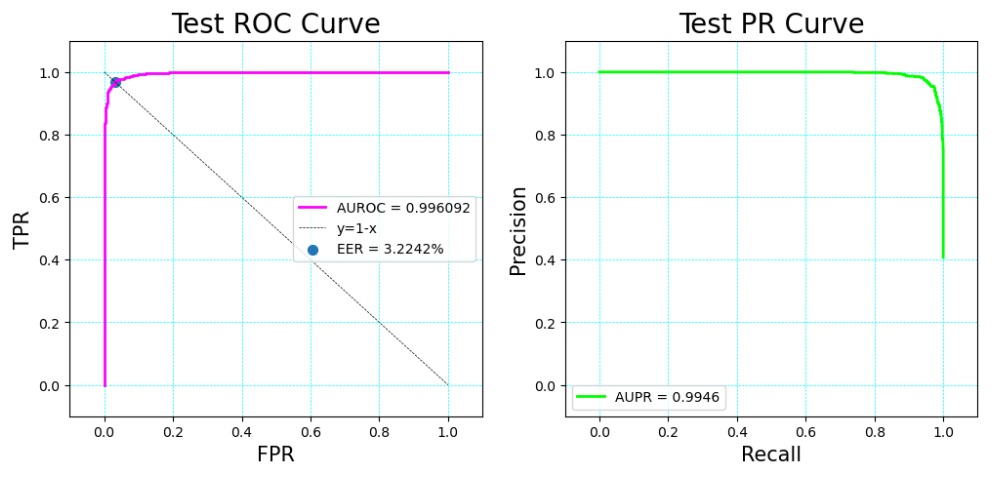} 
  \caption{(i)ROC and (ii)PR curve on the Testing set of  ICDAR SigComp Dutch dataset}
  \label{fig:roc_curve}
\end{figure}

\begin{figure}[ht]
  \centering
  \includegraphics[width=1\linewidth]{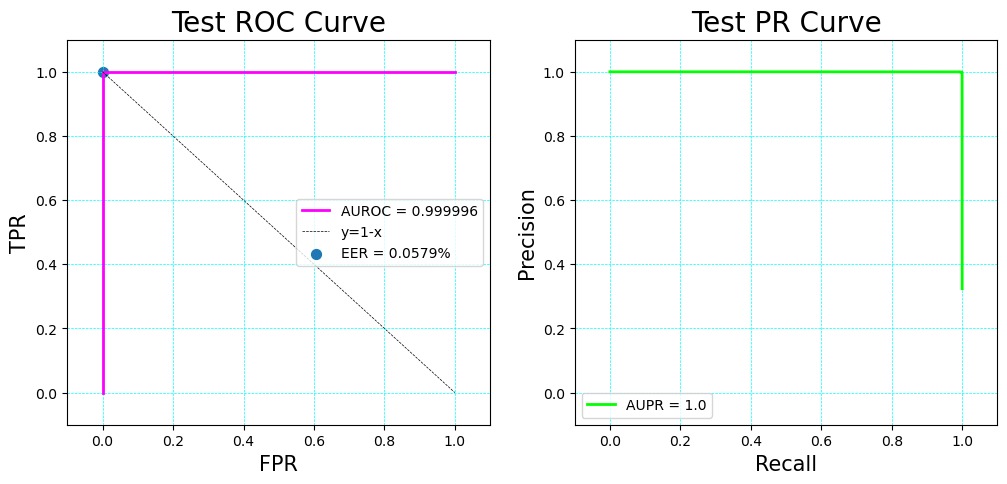} 
  \caption{(i)ROC and (ii)PR curve on the Testing set of  CEDAR dataset}
  \label{fig:pr_curve}
\end{figure}

We examined two histograms showing the Density vs. Euclidean distances and Count vs. Euclidean distances between embeddings to acquire additional insights into the model's performance. These histograms visually represent the distribution of distances between genuine pairings and imposter pairs, allowing us to assess the model's capability to distinguish between authentic and forged signatures based on the estimated embeddings. An important factor in successfully categorizing signatures is the optimal threshold. 

\begin{figure}[ht]
  \centering
  \includegraphics[width=1\linewidth]{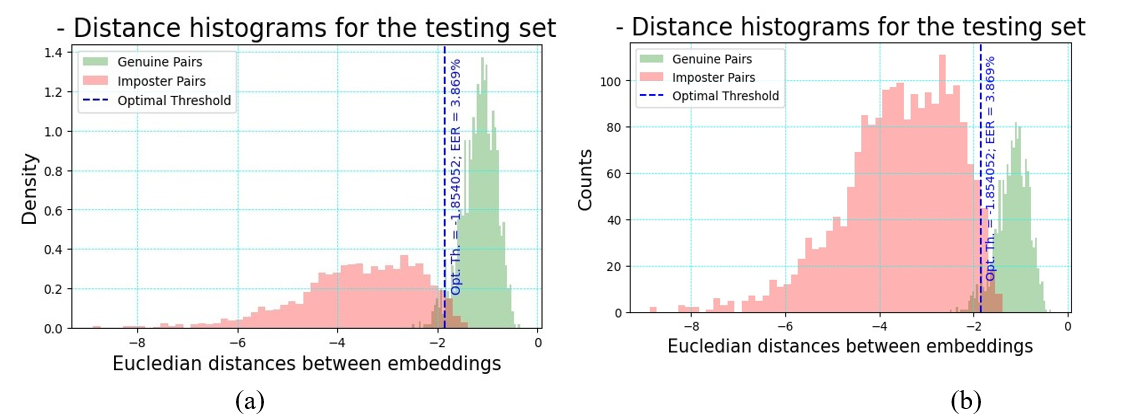} 
  \caption{Distance histograms for the Testing set on the ICDAR SigComp Dutch dataset}
  \label{fig:distance_histogram_icdar}
\end{figure}

\begin{figure}[ht]
  \centering
  \includegraphics[width=1\linewidth]{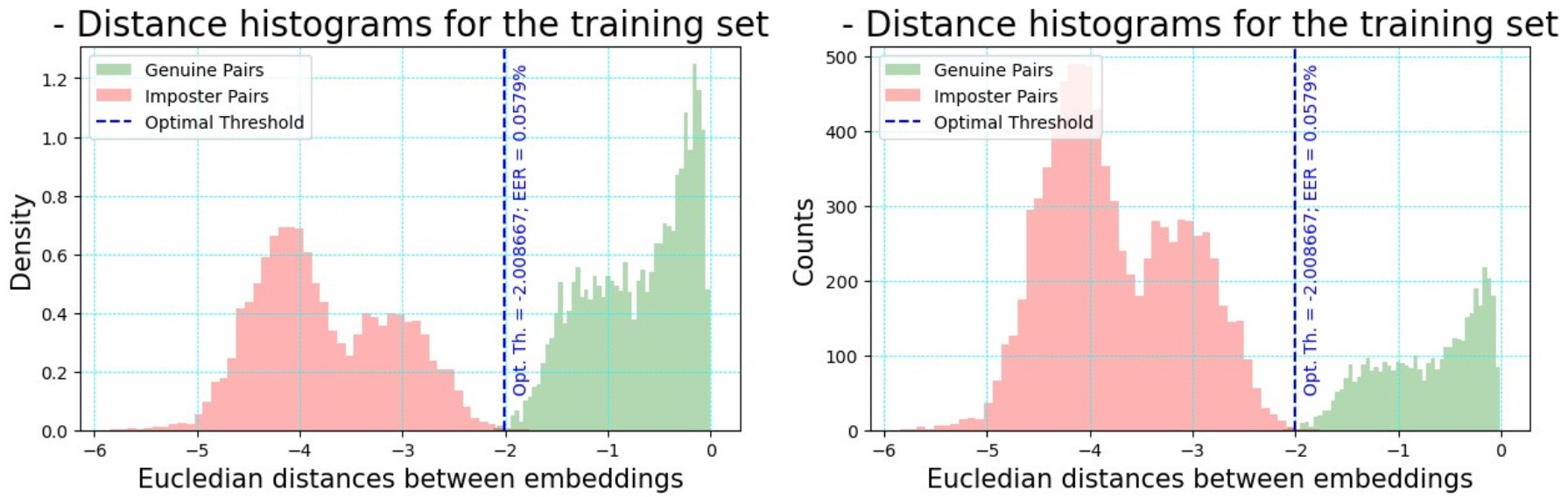} 
  \caption{Distance histograms for the Testing set on the CEDAR dataset}
  \label{fig:distance_histogram_cedar}
\end{figure}

Fig. \ref{fig:distance_histogram_icdar} and Fig. \ref{fig:distance_histogram_cedar} showcase the distance histograms for the Testing sets on the ICDAR SigComp Dutch dataset and the CEDAR dataset, respectively.

The False Match Rate (FMR) and False Non-Match Rate (FNMR) curve, which depicts the relationship between the error rates and the threshold value, provided another tool we used to assess the model's effectiveness. The Equal Error Rate (EER), a crucial measure in the detection of signature fraud, is represented at the junction of these two curves. In our instance, the EER at the threshold coordinate came out to be 3.22\% for the SigComp Dutch dataset and 0.058\% for the CEDAR dataset. This balance between false match and false non-match errors denotes the point at which the model performs at its best.

\begin{figure}[ht]
  \centering
  \includegraphics[width=0.8\linewidth]{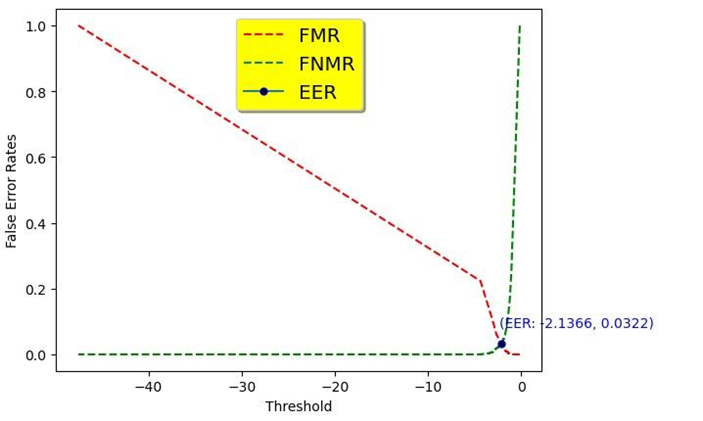} 
  \caption{False Match Rate and False Non-Match Rate Plot with EER representation for the ICDAR SigComp Dutch dataset}
  \label{fig:fmr_fnmr_icdar}
\end{figure}

\begin{figure}[ht]
  \centering
  \includegraphics[width=0.8\linewidth]{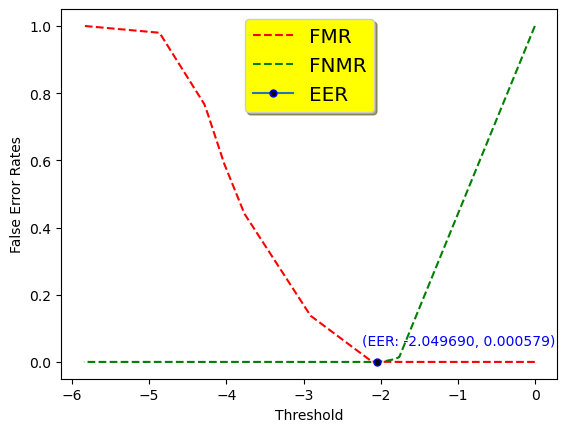} 
  \caption{False Match Rate and False Non-Match Rate Plot with EER representation for the CEDAR dataset}
  \label{fig:fmr_fnmr_cedar}
\end{figure}

Fig. \ref{fig:fmr_fnmr_icdar} and Fig. \ref{fig:fmr_fnmr_cedar} showcase the FMR and FNMR plot with the EER representation for the ICDAR SigComp Dutch dataset and the CEDAR dataset, respectively.

These assessment metrics—the ROC and PR curves, distance histograms, and the False Error Rate versus Threshold curve—all demonstrate the effectiveness of our Siamese model in detecting signature forgeries. The model's capacity to distinguish between authentic and forged signatures is supported by its strong AUC and AUPR values, and the distance histograms draw attention to the distinct differences between the two groups.

A quantitative evaluation of the model's performance at the best operating point is also provided by the EER computation. These results demonstrate the accuracy and dependability of the SigScatNet model in addressing the challenges related to signature forgery detection.

\section{Comparison with Other Models}
To evaluate the efficacy of our SigScatNet model, it is imperative to compare its performance against other state-of-the-art verification models in the research community, providing valuable insights into the strengths and weaknesses of our model and highlighting the advancements achieved by our proposed method. The results of these comparisons are presented in Table I and Table II.

\begin{table*}[ht]
\begin{adjustwidth}{-1in}{-1in} 

  \centering
  \caption{Comparison of proposed SigScatNet with other methods on the SigComp Dutch Dataset used in the research community}
  \label{tab:comparison_icdar}
  \begin{tabular}{|l|c|c|c|c|}
    \hline
    Method Used & Feature Extraction  & Classifier & Accuracy & EER \\
    \hline
    Cozzens et al. (2018) ~\cite{10} & CNN & CNN & 84.74 & - \\
    \hline
    Pham et al. (2014) ~\cite{11} & Global Features & Distance Based & 87.80  & - \\
    \hline
    Khan et al. (2015) ~\cite{12} & Local and Global Features & DTW & - & 5.27\% \\
    \hline
    Alvarez et al. (2016) ~\cite{13} & CNN & VGG-16 & 97.00 & - \\    \hline
    Chen et al. (2018) ~\cite{14} & Correlation Coefficients & Score based Likelihood & 96.17 & - \\    \hline
    Parodi and Gomez (2014) ~\cite{15} & - & Random Forest & - & 5.5\% \\    \hline
    \textbf{Proposed SigScatNet} & \textbf{Scattering + Siamese CNN} & \textbf{Distance Based} & \textbf{97.76} & \textbf{3.22\%} \\    \hline
  \end{tabular}
\end{adjustwidth}

\end{table*}

\begin{table*}[ht]
\begin{adjustwidth}{-1in}{-1in} 
\centering
\caption{Comparison of proposed SigScatNet with other methods on the CEDAR Dataset used in the research community}
\label{tab:comparison_cedar}
\begin{tabular}{|l|c|c|c|c|}
\hline
Method Used & Feature Extraction & Classifier & Accuracy & EER \\
\hline
Zois et al. ~\cite{16} & K-SVD & SVM & 84.74 & 2.78\% \\
\hline
Hadjadji et al. ~\cite{17} & Curvelet Transform & Bi-SVM, OC-PCA & 97.99 & - \\
\hline
Çalik et al. ~\cite{18} & CNN & 1-NN, CNN & 98.3 & - \\
\hline
Okawa ~\cite{19} & VLAD with KAZE & SVM & - & 1.0\% \\
\hline
Rateria et al. ~\cite{20} & CNN & CNN - SVM & 99.87 & - \\
\hline
Zois et al. ~\cite{21} & Poset Grid Features & SVM & - & 3.02\% \\
\hline
Eskander et al. ~\cite{22} & Feature learning & SVM & - & 4.63\% \\    \hline
Kumar et al. ~\cite{23} & Surrounding pixels & RBF-SVM & 91.67 & - \\    
\hline
Hafemann et al. ~\cite{24} & CNN (Signet-SPP) & SVM & - & 2.33\% \\    \hline
\textbf{Proposed SigScatNet} & \textbf{Scattering + Siamese CNN} & \textbf{Distance Based} & \textbf{99.91} & \textbf{0.058\%} \\   
\hline
\end{tabular}
\end{adjustwidth}
\end{table*}

Another important comparison to be done was to compare the SigScatNet model in two ways: one with Scattering parameters and one without Scattering parameters. Table III showcases the advantages of using the Scattering parameters.

\begin{table*}[ht]
  \centering
  \caption{Comparison of proposed SigScatNet with and without Scattering parameters on SigComp Dutch and on CEDAR Dataset}
  \label{tab:comparison_scattering}
  \begin{tabular}{|l|c|c|c|}
    \hline
    Methodology Used & Number of Parameters & EER (For SigComp) & EER (For CEDAR) \\
    \hline
    Only Siamese Network & 3,327,056 & \textbf{3.22\%} & 0.069\% \\
    \hline
    SigScatNet (Siamese CNN + Scattering) & \textbf{257,232} & 3.87\% & \textbf{0.058\%} \\
    \hline
  \end{tabular}
\end{table*}

The number of parameters in the plain Siamese network rose to around 3.3 million as illustrated in Table III to give comparable results. On the other hand, while incorporating the Scattering parameters in the network, we were able to reduce the parameters down to 257,232, which is a 92.27\% decrease in the number of parameters, still yielding a comparable or even better EER. This further showcases that incorporating wavelets into the network drastically reduces the complexity and dramatically increases computational efficiency.

\section{Conclusion and Future Work}
This research paper presents SigScatNet: a comprehensive and effective solution for signature forgery detection and similarity assessment by combining a Siamese deep learning network with Scattering wavelets. This system proves to be a reliable and effective model that can accurately differentiate between authentic and fake signatures while also offering a measure of similarity by utilizing the strength of deep learning and the distinctive characteristics of scattering wavelets. Furthermore, the addition of Scattering wavelets to our model helped us achieve excellent efficiency without sacrificing accuracy. In order to help the model make wise decisions, the wavelet-based feature extraction procedure extracted crucial structural and textural data from the signatures. The experimental evaluation of the SigComp Dutch and CEDAR datasets proved the viability of our suggested methodology. The SigScatNet model demonstrated excellent performance with an Equal Error Rate of 3.689\% for the SigComp Dutch Dataset and 0.05787\% for the CEDAR dataset after being trained on a wide range of real and fake signatures. This demonstrates the model's capacity to balance false match and false non-match errors, resulting in reliable and accurate detection of signature fraud.

The proposed approach presents a practical solution to address the challenges associated with signature forgery. Our model illustrates its potential for real-world deployment, enabling individuals and organizations to verify signatures quickly and accurately. It does this by utilizing cutting-edge deep-learning techniques and harnessing the special properties of scattering wavelets. By offering a strong and effective framework that combines the strength of Siamese deep learning networks and Scattering wavelets, this research makes a contribution to the field of signature fraud detection and paves the way for increased security and confidence in applications for signature verification. Since this model is extremely lightweight, it can in theory be implemented into an app that can make the signature verification task in the palm of everyone’s hands.

All other authors have no conflicts of interest. SigComp Dutch dataset analyzed during the current study is available in the ICDAR 2011 Signature Verification Competition (SigComp2011) repository. CEDAR dataset analyzed during the current study is available in the CEDAR from University of Buffalo repository.

\bibliographystyle{unsrt}
 \bibliography{conference_101719}

\end{document}